\pdfoutput=1
\documentclass[letterpaper]{article}
\usepackage[preprint]{paperstyle}
\usepackage[utf8]{inputenc}
\usepackage[hyphens]{url}
\usepackage{graphicx}
\usepackage{natbib}
\usepackage{caption}
\usepackage{amsmath}
\usepackage{amssymb}
\usepackage{booktabs}
\usepackage{float}
\usepackage{tcolorbox}
\title{CHIME: Credit-Aware Hierarchical Memory Evolution for\\Long-Horizon Agentic Planning}

\author{
Yongshi Ye\textsuperscript{1,2,4},
Tian Lan\textsuperscript{4},
Feihu Jiang\textsuperscript{4},
Muyang Ye\textsuperscript{3},
Bin Zhu\textsuperscript{4},\\
Qianghuai Jia\textsuperscript{4},
Longyue Wang\textsuperscript{4}\corresponding,
Zhao Xu\textsuperscript{4},
Weihua Luo\textsuperscript{4},
Xiaodong Shi\textsuperscript{1,2}\corresponding
}
\affiliations{
\textsuperscript{1}Institute of Artificial Intelligence, Xiamen University\\
\textsuperscript{2}Key Laboratory of Digital Protection and Intelligent Processing of Intangible Cultural Heritage\\
of Fujian and Taiwan (Xiamen University), Ministry of Culture and Tourism\\
\textsuperscript{3}Zhejiang University\\
\textsuperscript{4}Alibaba Group
}

\date{}

\begin{document}

\maketitle

\begin{abstract}
Planning is a central capability that enables agents to decompose complex
long-horizon tasks into manageable steps.
Test-time search and training-based methods improve planning but incur high
inference costs or require expensive training data.
Self-evolving memory instead accumulates reusable experience from agent
interaction outcomes into an external memory bank, so planning capability
keeps improving at inference time without parameter updates.
However, existing self-evolving memory methods share an inherent
credit assignment problem: they rely on final task outcomes as feedback, but
such outcomes conflate plan quality with execution errors and environmental
factors, so the accumulated planning experience is often biased and noisy.
To address this problem, we propose \textbf{C}redit-Aware \textbf{HI}erarchical
\textbf{M}emory \textbf{E}volution (CHIME), a self-evolving memory framework
that maintains a separate planning bank and execution bank and follows an
\emph{attribute-before-memorize} principle: CHIME first attributes each task outcome to the plan, the execution, both, or neither, and then updates only the corresponding memory bank.
Extensive experiments on four long-horizon agent benchmarks show that CHIME consistently outperforms state-of-the-art training-based and
self-evolving memory baselines.
Further analyses reveal several interesting findings.
For example, CHIME accumulates effective memory with far
fewer items.
In addition, the learned memory values faithfully reflect downstream utility: high-quality planning memories are more valuable than execution memories.
Finally, the accumulated memory effectively transfers across backbone models.
Code will be released at \url{https://github.com/ATH-MaaS/Marco-DeepResearch}.
\end{abstract}

\section{Introduction}

Long-horizon tasks pose a defining challenge for agentic systems: agents 
must coordinate decisions across interdependent steps while sustaining 
adherence to task constraints throughout execution~\cite{xie2024travelplannerbenchmarkrealworldplanning,wang2024describeexplainplanselect,erdogan2025planandactimprovingplanningagents}. 
Agentic planning addresses these requirements by decomposing the overall objective into modular, independently executable subgoals~\cite{wang2024describeexplainplanselect,shen2023hugginggpt,zhang2026deepplanningbenchmarkinglonghorizonagentic,liu2026todoevolvelearningarchitectagent} 
and establishing the dependencies and constraints among them~\cite{kim2024llmcompilerparallelfunction,guo2024castlconstraintsspecificationsllm,lan-etal-2026-table}, 
making it a central capability for solving long-horizon tasks.

This has motivated prior work to improve agents' planning capability.
Test-time search~\cite{yao2023treethoughtsdeliberateproblem,hao2023reasoninglanguagemodelplanning,zhou2024languageagenttreesearch,yu2026webanchoranchoringagentplanning,coreteam2026mimov2flashtechnicalreport}
improves per-task performance by exploring multiple candidate plans or trajectories,
but incurs high inference costs and does not retain reusable experience.
To retain and reuse planning experience across tasks, subsequent work has
pursued two directions:
(1) Planner model training~\cite{erdogan2025planandactimprovingplanningagents,si2026goalplanjustwish,liu2026todoevolvelearningarchitectagent,miromindteam2026mirothinker17h1heavyduty}
internalizes planning capability into model parameters, but requires costly
data collection and post-training, since identifying high-quality plans
relies on extensive downstream executions;
(2) Self-evolving memory~\cite{chen2026amapreduceexecutingwidesearch,ouyang2026reasoningbankscalingagentselfevolving}
instead accumulates experience in an external, training-free memory bank,
offering a more efficient and scalable alternative.
We therefore focus on improving planning within the self-evolving memory
paradigm in this paper.
\begin{figure}[t]
    \centering
    \includegraphics[width=\linewidth]{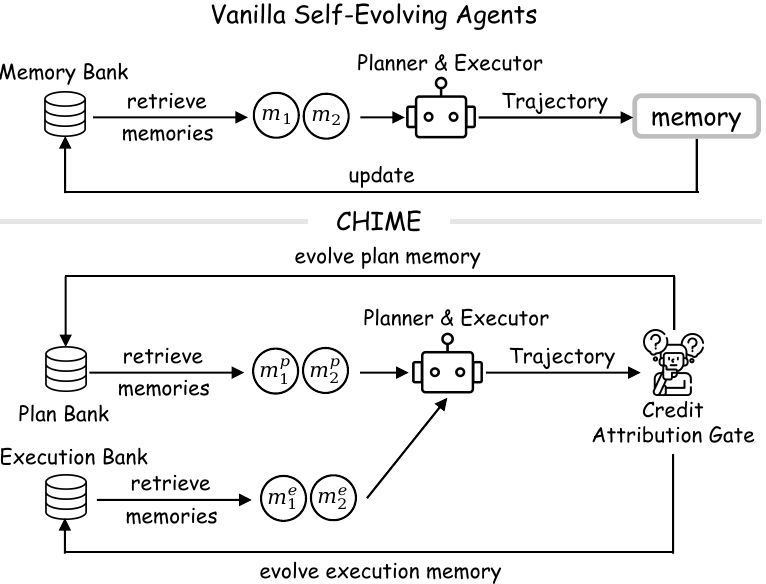}
    \caption{\textbf{Vanilla self-evolving agents vs. CHIME.} Vanilla self-evolving agents write experience directly to shared memory. CHIME attributes task outcomes through a credit attribution gate before updating planning or execution memory.}
    \label{fig:chime_overview}
\end{figure}

However, existing self-evolving memory methods share a fundamental
limitation for agentic planning: they equate downstream task outcomes with plan quality.
The same outcome can arise from different causes: success may result
from a sound plan or from executor recovery from a flawed one, while
failure may stem from a flawed plan, incorrect execution, or
environmental issues.
Using such outcomes directly as memory feedback therefore leads to a
fundamental credit assignment failure.
On the one hand, once biased experience is written into the memory
bank, it is repeatedly retrieved and reused across tasks, allowing
local misjudgments to accumulate into systematic planning errors.
On the other hand, execution- or environment-level
experience may be mistaken for planning memory;
retrieving might low-level details then misguides high-level planning decisions instead of providing planning guidance.

To address this, we propose \textbf{C}redit-Aware 
\textbf{HI}erarchical \textbf{M}emory \textbf{E}volution (CHIME), a 
self-evolving agentic planning framework that replaces outcome-based memory updates~\cite{ye2026umemunifiedmemoryextraction} with an \emph{attribute-before-memorize} process. 
Specifically, CHIME comprises three components:
(1) \textbf{Hierarchical Memory Bank} explicitly separates 
planning and execution memory into a \emph{planning bank} and an \emph{execution bank}, retrieving stage-specific memories through similarity retrieval and value-based reranking to separately guide the agent's planning and execution;
(2) \textbf{Credit Attribution Gate} then reflects on the task, plan, execution, and retrieved memories to attribute reusable feedback to planning, execution, both stages, or neither, before any memory is written; it also generates stage-specific experience and identifies 
misleading memories;
and (3) \textbf{Credit-Aware Memory Evolution} updates the values of retrieved memories using confidence-weighted feedback, while 
filtering, merging, or inserting new experience only into the 
attributed bank. 
In this way, CHIME evolves each bank only with memory attributed 
to that stage, preventing biased signals from persisting and 
execution-level memory from contaminating the planning bank.

Experiments on four long-horizon benchmarks with two
backbone models show that CHIME consistently outperforms training-based
and self-evolving memory baselines, improving the eval average over
the strongest baseline by 2.96\% and 3.68\% on the two
backbones, respectively.
Further analyses reveal several interesting findings:
(1)~CHIME achieves the highest accuracy while retaining only
129 memories, versus 3,585 for the strong baseline;
(2)~the learned memory values faithfully reflect downstream
utility, and planning memories bring more than twice the
accuracy gain of execution ones (21.7\%$\to$50.8\% vs.\
23.7\%$\to$35.4\%);
(3)~the Credit Attribution Gate is reliable, with repeated
attributions agreeing in up to 97.9\% of cases and over half
of the failures rescued by the generated experience;
(4)~replacing model's self-reflection credit with a stronger credit model yields a further gain of up to 3.12\%;
and (5)~the accumulated memory effectively transfers across backbones, outperforming A-MapReduce's transferred memory by up to 4.68\%.
These findings confirm that attribute-before-memorize
is the key to CHIME's effectiveness.

\section{Related Work}

\paragraph{Self-evolving memory.}
Self-evolving memory methods treat an external memory bank as
the agent's evolvable parameters, distilling interaction
trajectories into reusable experience or
skills~\cite{yu2025memagentreshapinglongcontext,wang2025memalphalearningmemory,zhang2026memrlselfevolvingagents}.
Unlike training-based methods, they keep model parameters
frozen and update only the memory bank, avoiding costly
trajectory collection and
post-training~\cite{li2025deepagentgeneralreasoning,wu2025evolverselfevolvingagents}.
For example,
ReasoningBank~\cite{ouyang2026reasoningbankscalingagentselfevolving}
distills generalizable reasoning strategies from self-judged
successful and failed trajectories, and
UMEM~\cite{ye2026umemunifiedmemoryextraction} jointly optimizes
memory extraction and management with reinforcement learning.

\paragraph{Agentic planning.}
Agentic planning decomposes complex objectives into structured
steps and coordinates their execution.
It underlies a broad range of agent
systems~\cite{yao2023reactsynergizingreasoningacting,erdogan2025planandactimprovingplanningagents,liu2026todoevolvelearningarchitectagent,lan-etal-2026-table}.
Long-horizon tasks often involve many interdependent subtasks. Without an explicit plan to organize them, agents easily lose
track of progress, a failure known as the lost-in-the-middle
problem~\cite{lan-etal-2026-table,liu2026todoevolvelearningarchitectagent}.
As recent benchmarks introduce increasingly challenging
tasks, improving planning capability has become a central research problem for agent systems~\cite{wong2025widesearchbenchmarkingagenticbroad,lan2025deepwidesearchbenchmarkingdepthwidth,yang2025hscodecomprealisticexpertlevel}.

\paragraph{Improving agentic planning.}
Existing efforts fall into three paradigms:
(1) \textbf{Test-time search} explores multiple candidate plans
or trajectories at inference time and selects among them with
value estimates or environmental
feedback~\cite{yao2023treethoughtsdeliberateproblem,hao2023reasoninglanguagemodelplanning,zhou2024languageagenttreesearch}.
For example,
MiroThinker-H1~\cite{miromindteam2026mirothinker17h1heavyduty}
and WebAnchor~\cite{yu2026webanchoranchoringagentplanning}
select high-quality plans via rejection sampling;
(2) \textbf{Planner model training} internalizes planning
capability into model
parameters~\cite{erdogan2025planandactimprovingplanningagents,si2026goalplanjustwish,yu2026webanchoranchoringagentplanning}.
For example, TodoEvolve~\cite{liu2026todoevolvelearningarchitectagent}
constructs a modular design space of planning architectures and
trains a meta-planner with reinforcement learning to synthesize
task-specific planning systems.
However, such methods require costly trajectory collection and
post-training, and the resulting planner is expensive to update
continually~\cite{liu2026todoevolvelearningarchitectagent}; and
(3) \textbf{Self-evolving memory} offers a
training-free and continual alternative, but few works evolve memory
specifically for
agentic planning~\cite{kagaya2024rapretrievalaugmentedplanningcontextual}.
A representative is
A-MapReduce~\cite{chen2026amapreduceexecutingwidesearch}, which
evolves structured hints from past executions to improve task
decomposition and result aggregation in long-horizon agentic
search.
However, all three paradigms evaluate agent's plans by final task outcomes, implicitly equating outcome with plan quality.
This equivalence does not hold, as outcomes are also shaped by
execution details and environmental factors.

\section{Methodology of CHIME}
\label{sec:method}

\subsection{Task Formulation of Self-Evolving Agents}
\label{sec:task_formulation}

We consider a self-evolving agent that solves a stream of tasks.
The agent's policy parameters remain frozen, and adaptation is
carried by an external memory bank $\mathcal{M}_t$ that evolves
across episodes.
At episode $t$, the agent receives a task $x_t$ and retrieves
task-relevant experience $\mathcal{M}_t^{\mathrm{ret}}$ from
$\mathcal{M}_t$.
Conditioned on $x_t$ and $\mathcal{M}_t^{\mathrm{ret}}$, the frozen
policy produces a trajectory $\tau_t$ and receives an outcome $s_t$
from the environment and the evaluator.
A memory update operator then distills the episode into reusable
experience:
\begin{equation}
    \mathcal{M}_{t+1}
    =
    \mathcal{U}
    \left(
    \mathcal{M}_t,x_t,\tau_t,s_t
    \right).
    \label{eq:self_evolve}
\end{equation}
During evaluation, the memory bank is frozen, so that performance
reflects experience accumulated before evaluation.
Existing self-evolving memory methods mainly use the final outcome
$s_t$ as the supervision signal for memory updates.
For agentic planning, this signal is biased: the outcome
depends on both the plan and the execution process, so it
does not faithfully reflect plan quality.

\subsection{Overview of CHIME}
\label{sec:method_overview}

\begin{figure*}[t]
    \centering
    \includegraphics[width=\linewidth]{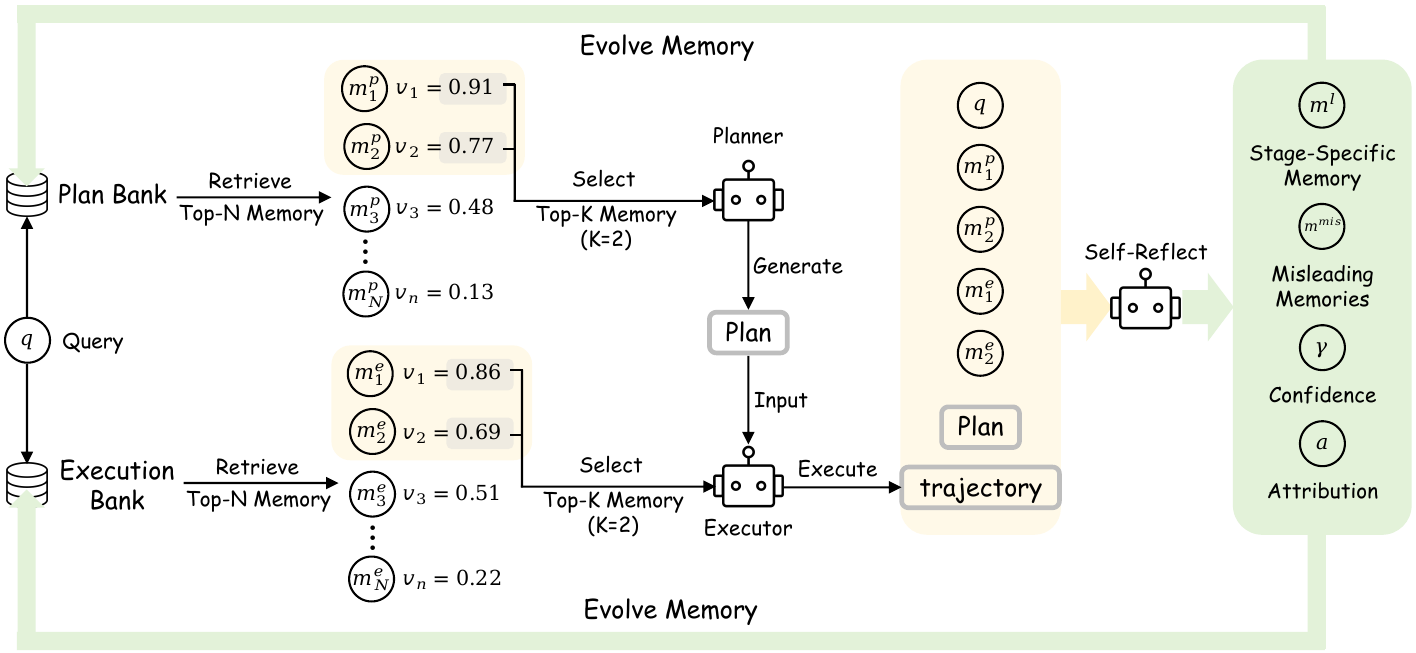}
    \caption{\textbf{Overview of CHIME.} \textbf{Left:} the Hierarchical Memory Bank maintains the planning and execution memory banks. For each task, similarity retrieval and value-based reranking select stage-specific memories to guide the planner and the executor. \textbf{Right:} at each episode, the Credit Attribution Gate reflects on the task, plan, execution, and retrieved memories, attributes reusable feedback to planning, execution, both, or neither, and generates stage-specific experience. \textbf{Feedback loop:} Credit-Aware Memory Evolution updates the memory values, and filters, merges, or inserts new experience into the attributed bank.}
    \label{fig:chime}
\end{figure*}

We propose CHIME (Figure~\ref{fig:chime}), an
attribute-before-memorize self-evolving framework for agentic
planning.
It attributes each task outcome to planning, execution, or external
factors, and updates only the corresponding memory bank, which
reduces biased feedback and cross-stage memory contamination.
CHIME consists of three components: a Hierarchical Memory Bank
that separates planning and execution experience, a Credit
Attribution Gate that attributes each outcome to its responsible
stage, and Credit-Aware Memory Evolution that updates the banks
from the attributed credit.

\paragraph{Hierarchical Memory Bank.}

Unlike existing self-evolving
methods~\cite{ye2026umemunifiedmemoryextraction}, CHIME maintains
hierarchical memory banks
\begin{equation}
    \mathcal{M}_t=
    \left(
    \mathcal{M}_{\mathrm{plan},t},
    \mathcal{M}_{\mathrm{exec},t}
    \right),
\end{equation}
which comprises a planning bank and an execution bank, indexed by
$\ell\in\{\mathrm{plan},\mathrm{exec}\}$.
The planning bank stores strategic experience (e.g., subtask
decomposition, dependency modeling, and constraint
incorporation), whereas the execution bank stores operational
experience (e.g., tool selection and invocation).
Each bank consists of memory items
\begin{equation}
    m_i=(c_i,e_i,v_i,n_i),
    \label{eq:memory_item}
\end{equation}
where $c_i$ describes the tasks to which the memory applies,
serving as the key for memory retrieval and merging;
$e_i$ is the reusable experience memory distilled from past
episodes;
$v_i\in[-1,1]$ is a value score estimated from reuse feedback,
which guides reranking (Eq.~\ref{eq:reranking}) and pruning;
and $n_i$ is the reuse count, which calibrates the value score
(Eq.~\ref{eq:adjusted_value}).
At episode $t$, CHIME receives a long-horizon task $x_t$ and
constructs a planning query $q_{\mathrm{plan},t}$ to retrieve
$\mathcal{M}_{\mathrm{plan},t}^{\mathrm{ret}}$ from
$\mathcal{M}_{\mathrm{plan},t}$.
The planner then generates a plan $p_t$ from $x_t$ and
$\mathcal{M}_{\mathrm{plan},t}^{\mathrm{ret}}$.
Next, CHIME constructs an execution query $q_{\mathrm{exec},t}$ from
$(x_t,p_t)$ and retrieves
$\mathcal{M}_{\mathrm{exec},t}^{\mathrm{ret}}$
from $\mathcal{M}_{\mathrm{exec},t}$.
The executor uses $p_t$ and $\mathcal{M}_{\mathrm{exec},t}^{\mathrm{ret}}$
to produce a trajectory $\tau_t$, and the environment returns a
binary outcome $s_t\in\{0,1\}$ indicating task success.
Finally, the credit attribution gate produces a diagnosis $g_t$
from the plan, trajectory, outcome, and retrieved memories, which
drives the transition from $\mathcal{M}_t$ to $\mathcal{M}_{t+1}$.
One CHIME episode is summarized as
\begin{equation}
    \mathcal{M}_t
    \!\rightarrow\!
    \mathcal{M}_{\mathrm{plan},t}^{\mathrm{ret}}
    \!\rightarrow\!
    p_t
    \!\rightarrow\!
    \mathcal{M}_{\mathrm{exec},t}^{\mathrm{ret}}
    \!\rightarrow\!
    \tau_t
    \!\rightarrow\!
    s_t
    \!\rightarrow\!
    g_t
    \!\rightarrow\!
    \mathcal{M}_{t+1}.
\end{equation}
These steps correspond to the left, right, and feedback-loop parts
of Figure~\ref{fig:chime}.

\paragraph{Hierarchical Memory Retrieval.}

Retrieval operates independently at each stage
$\ell\in\{\mathrm{plan},\mathrm{exec}\}$.
Given the stage-specific query $q_{\ell,t}$, CHIME retrieves the
Top-$N$ memories from the corresponding bank $\mathcal{M}_{\ell,t}$
according to $\mathrm{sim}(q_{\ell,t},c_i)$, yielding the candidate
set $\mathcal{C}_{\ell,t}$.
For each candidate memory $m_i\in\mathcal{C}_{\ell,t}$, CHIME
adjusts its value score according to its reuse count:
$\widetilde{v}_i=v_i\frac{n_i}{n_i+\lambda}\label{eq:adjusted_value}$,
where $n_i$ is the reuse count of $m_i$, and $\lambda$ is a
smoothing constant.
The factor shrinks the value estimates of rarely reused memories
toward zero, so that they fall back to similarity-only ranking.
Semantic similarity does not indicate whether a memory actually
helps, so CHIME reranks the candidates by their credited value and
retrieves the Top-$K$:
\begin{equation}
    \mathcal{M}_{\ell,t}^{\mathrm{ret}}
    =
    \underset{m_i\in\mathcal{C}_{\ell,t}}{\mathrm{TopK}}\;
    \mathrm{sim}(q_{\ell,t},c_i)
    \left[
    1+\mathrm{clip}
    \left(
    \alpha\widetilde{v}_i,-\beta,\beta
    \right)
    \right],
    \label{eq:reranking}
\end{equation}
where $\alpha$ is the value weight, and $\beta$ bounds the value
effect so that similarity remains the dominant ranking factor.
As a result, memories validated as helpful in past episodes are
preferred, whereas misleading ones are suppressed.
The retrieved memories guide the corresponding planning or execution
stage.

\paragraph{Credit Attribution Gate.}

After planning and execution with the retrieved memories, the
agent obtains the task outcome $s_t$.
The credit attribution gate then attributes this outcome to
planning, execution, or external factors.
CHIME implements the gate as a structured self-reflection of the
frozen policy: the model is prompted to review the task, plan,
trajectory, outcome, and retrieved memories, assess plan
sufficiency, execution correctness, and the effects of external
factors, and produce a diagnosis
\begin{equation}
    g_t=
    \left(
    a_t,
    \{(c_{\ell,t}^{\mathrm{new}},e_{\ell,t}^{\mathrm{new}})\}_{\ell},
    \gamma_t,
    \mathcal{M}_t^{\mathrm{mis}}
    \right),
    \label{eq:gate}
\end{equation}
where the attribution label $a_t$ takes one of four values:
\emph{planning} (e.g., an inadequate plan), \emph{execution}
(e.g., incorrect execution of a sufficient plan), \emph{both}
(e.g., distinct issues at both stages), and \emph{none} (e.g.,
external causes, insufficient evidence, or no reusable experience);
in successful episodes, it instead identifies the stage credited
with the success;
$(c_{\ell,t}^{\mathrm{new}},e_{\ell,t}^{\mathrm{new}})$ is the task
scenario and experience generated for each stage
$\ell\in\{\mathrm{plan},\mathrm{exec}\}$;
$\gamma_t\in[0,1]$ is the attribution confidence;
and $\mathcal{M}_t^{\mathrm{mis}}$ contains retrieved memories
identified as misleading.
These outputs determine the subsequent memory value and content
updates.

\paragraph{Credit-Aware Memory Evolution.}

CHIME then uses the diagnosis $g_t$ to evolve the memory bank in
two steps, both directed at the attributed stages:
(1) \textbf{~\emph{Memory value evolution}}:
CHIME evaluates each retrieved memory $m_i$ with a value feedback
$\delta_{i,t}$: the memory is rewarded when its stage is credited
for a successful outcome, penalized when it is identified as
misleading, and unchanged otherwise.
The attribution label determines the memory stages to update:
\begin{equation}
\mathcal{L}(a_t)=
\begin{cases}
\{\mathrm{plan}\}, & a_t=\mathrm{planning},\\
\{\mathrm{exec}\}, & a_t=\mathrm{execution},\\
\{\mathrm{plan},\mathrm{exec}\}, & a_t=\mathrm{both},\\
\emptyset, & a_t=\mathrm{none}.
\end{cases}
\label{eq:attributed_stages}
\end{equation}
The feedback for memory $m_i$ in stage $\ell(i)$ is then
\begin{equation}
    \delta_{i,t}=
    \begin{cases}
        -\gamma_t, & m_i\in\mathcal{M}_t^{\mathrm{mis}},\\
        +\gamma_t, & s_t=1 \text{ and } \ell(i)\in\mathcal{L}(a_t),\\
        0, & \text{otherwise},
    \end{cases}
    \label{eq:value_feedback}
\end{equation}
The feedback magnitude is the attribution confidence $\gamma_t$.
CHIME updates the value score and reuse count using an online average:
\begin{equation}
    v_i\leftarrow
    \mathrm{clip}\!\left(
    \frac{n_i v_i+\delta_{i,t}}{n_i+1},-1,1
    \right), \qquad
    n_i\leftarrow n_i+1.
    \label{eq:value_update}
\end{equation}
The updated $v_i$ affects subsequent retrieval through Eq.~\ref{eq:reranking}.
To maintain the quality of the memory bank, CHIME removes a
memory when it has been reused sufficiently often yet remains
low-valued, i.e., $n_i\geq n_{\min}$ and
$v_i<\theta_{\mathrm{prune}}$.
(2)\textbf{~\emph{Memory content evolution}}:
For each attributed stage $\ell\in\mathcal{L}(a_t)$, CHIME
updates the corresponding bank with the new experience in one of three
ways: merging it into a sufficiently similar memory, inserting it as a new memory item, or directly discarding it.
\begin{equation}
\begin{cases}
\text{merge } e_{\ell,t}^{\mathrm{new}} \text{ into } m_{i^*},
    & \text{if accepted and similar},\\
\text{insert } (c_{\ell,t}^{\mathrm{new}},e_{\ell,t}^{\mathrm{new}},0,0),
    & \text{if accepted but not similar},\\
\text{discard the experience},
    & \text{otherwise},
\end{cases}
\label{eq:content_update}
\end{equation}
where $m_{i^*}$ is the most similar memory in the corresponding memory bank,
with similarity defined as
$\mathrm{sim}(c_{\ell,t}^{\mathrm{new}},c_{i^*})>\theta_{\mathrm{merge}}$;
acceptance requires $\gamma_t\geq\theta_{\mathrm{conf}}$ and a
concrete task scenario with actionable content.
A newly inserted memory starts with a neutral value and zero
reuse count, so its value is estimated from future reuse rather
than inherited from the gate.
Applying this update only to the attributed stages yields
$\mathcal{M}_{t+1}$ and limits cross-stage memory contamination.
\section{Experimental Setup}
\label{sec:experimental_setup}

\begin{table*}[t]
\centering
\small
\renewcommand{\arraystretch}{1.08}
\begin{tabular*}{\textwidth}{@{\extracolsep{\fill}}lcccccccccc@{}}
\toprule
& \multicolumn{2}{c}{\textbf{$\tau^2$-Bench}}
& \multicolumn{2}{c}{\textbf{VitaBench}}
& \multicolumn{2}{c}{\textbf{BrowseComp-ZH}}
& \multicolumn{2}{c}{\textbf{BFCL-v4}}
& \multicolumn{2}{c}{\textbf{Average}} \\
\cmidrule(lr){2-3}\cmidrule(lr){4-5}\cmidrule(lr){6-7}\cmidrule(lr){8-9}\cmidrule(lr){10-11}
\textbf{Method}
& \textbf{Train} & \textbf{Eval}
& \textbf{Train} & \textbf{Eval}
& \textbf{Train} & \textbf{Eval}
& \textbf{Train} & \textbf{Eval}
& \textbf{Train} & \textbf{Eval} \\
\midrule

\multicolumn{11}{c}{\textbf{Qwen3.5-Flash}} \\
\midrule
Qwen3.5-Flash No-Plan
& 21.68 & 22.32
& 9.40 & 8.89
& 20.20 & 21.25
& 71.33 & 70.86
& 30.65 & 30.83 \\
WebAnchor~\citep{yu2026webanchoranchoringagentplanning}
& 24.77 & 23.45
& 14.76 & 11.94
& 17.68 & 21.98
& 69.36 & 70.26
& 31.64 & 31.91 \\
TodoEvolve~\citep{liu2026todoevolvelearningarchitectagent}
& 20.13 & 21.28
& 10.60 & 6.11
& 20.88 & 20.51
& 69.75 & 70.20
& 30.34 & 29.53 \\
A-MapReduce~\citep{chen2026amapreduceexecutingwidesearch}
& 19.81 & 20.55
& 12.02 & 6.67
& 18.35 & 18.31
& 73.84 & 72.45
& 31.01 & 29.50 \\
\textbf{CHIME} (Our Method)
& \textbf{26.25} & \textbf{25.92}
& \textbf{16.19} & \textbf{15.83}
& \textbf{23.74} & \textbf{23.81}
& \textbf{73.90} & \textbf{73.91}
& \textbf{35.02} & \textbf{34.87} \\

\midrule
\multicolumn{11}{c}{\textbf{DeepSeek-V4-Flash}} \\
\midrule
DeepSeek-V4-Flash No-Plan
& 31.42 & 32.86
& 16.31 & 11.11
& 26.94 & 23.81
& 63.75 & 64.15
& 34.61 & 32.98 \\
WebAnchor~\cite{yu2026webanchoranchoringagentplanning}
& 31.30 & 29.61
& 22.62 & 17.50
& 29.46 & 24.54
& 58.27 & 60.88
& 35.41 & 33.13 \\
TodoEvolve~\cite{liu2026todoevolvelearningarchitectagent}
& 28.45 & 26.18
& 18.57 & 15.56
& 26.26 & 22.34
& 56.29 & 58.32
& 32.39 & 30.60 \\
A-MapReduce~\cite{chen2026amapreduceexecutingwidesearch}
& 30.55 & 30.95
& 21.07 & 15.83
& 27.61 & 24.54
& \textbf{69.18} & \textbf{69.98}
& 37.10 & 35.33 \\
\textbf{CHIME} (Our Method)
& \textbf{33.15} & \textbf{36.28}
& \textbf{22.98} & \textbf{21.11}
& \textbf{29.97} & \textbf{31.50}
& 65.89 & 67.15
& \textbf{38.00} & \textbf{39.01} \\

\bottomrule
\end{tabular*}
\caption{Results (\%) averaged over three random instance orderings (Avg@3) across two backbone models and four benchmarks. Average is computed across benchmarks within each split. \textbf{Bold} indicates the best result in each column under each backbone.}
\label{tab:main_results}
\end{table*}

\begin{table*}[t]
\centering
\small
\renewcommand{\arraystretch}{1.08}
\begin{tabular*}{\textwidth}{@{\extracolsep{\fill}}lcccccccccc@{}}
\toprule
& \multicolumn{2}{c}{\textbf{$\tau^2$-Bench}}
& \multicolumn{2}{c}{\textbf{VitaBench}}
& \multicolumn{2}{c}{\textbf{BrowseComp-ZH}}
& \multicolumn{2}{c}{\textbf{BFCL-v4}}
& \multicolumn{2}{c}{\textbf{Average}} \\
\cmidrule(lr){2-3}\cmidrule(lr){4-5}\cmidrule(lr){6-7}\cmidrule(lr){8-9}\cmidrule(lr){10-11}
\textbf{Method}
& \textbf{Train} & \textbf{Eval}
& \textbf{Train} & \textbf{Eval}
& \textbf{Train} & \textbf{Eval}
& \textbf{Train} & \textbf{Eval}
& \textbf{Train} & \textbf{Eval} \\
\midrule
\textbf{CHIME}
& \textbf{26.25} & \textbf{25.92}
& 16.19 & \textbf{15.83}
& \textbf{23.74} & \textbf{23.81}
& \textbf{73.90} & \textbf{73.91}
& \textbf{35.02} & \textbf{34.87} \\
\midrule
\multicolumn{11}{l}{\textit{Hierarchical Memory Design}} \\
\quad w/o Plan Memory
& 22.96 & 23.11
& 15.95 & 12.22
& 20.71 & 20.51
& 69.52 & 70.60
& 32.29$_{\downarrow 2.7}$ & 31.61$_{\downarrow 3.3}$ \\
\quad w/o Execution Memory
& 24.38 & 25.14
& 15.47 & 12.50
& 20.88 & 9.16
& 72.93 & 73.42
& 33.42$_{\downarrow 1.6}$ & 30.06$_{\downarrow 4.8}$ \\
\quad w/o All Memory
& 24.05 & 23.67
& 13.22 & 12.50
& 19.70 & 19.78
& 69.52 & 70.99
& 31.62$_{\downarrow 3.4}$ & 31.74$_{\downarrow 3.1}$ \\
\quad w/o Hierarchical Bank
& 24.87 & 22.76
& 15.60 & 11.11
& 21.04 & 19.41
& 72.63 & 71.79
& 33.54$_{\downarrow 1.5}$ & 31.27$_{\downarrow 3.6}$ \\
\midrule
\multicolumn{11}{l}{\textit{Credit Attribution and Memory Evolution}} \\
\quad w/o Attribution Gate
& 21.21 & 22.28
& \textbf{17.02} & 12.50
& 20.20 & 22.71
& 73.40 & 72.23
& 32.96$_{\downarrow 2.1}$ & 32.43$_{\downarrow 2.4}$ \\
\quad w/o Credit Rerank
& 24.32 & 24.75
& 16.31 & 10.83
& 21.72 & 23.44
& 73.75 & 72.28
& 34.03$_{\downarrow 1.0}$ & 32.83$_{\downarrow 2.0}$ \\
\quad w/o Memory Evolution
& 25.61 & 24.32
& 16.79 & 8.61
& 20.71 & 22.34
& 73.79 & 72.05
& 34.23$_{\downarrow 0.8}$ & 31.83$_{\downarrow 3.0}$ \\
\bottomrule
\end{tabular*}
\caption{Ablation study results (\%) with Qwen3.5-Flash, averaged over three random orderings (Avg@3). Average is computed across benchmarks within each split; downward values indicate drops from CHIME. Best results are \textbf{bolded}.}
\label{tab:ablation_results}
\end{table*}

\paragraph{Baselines.}
We compare CHIME with four baselines, ordered by increasing proximity to our setting:
(1)~\textbf{No-Plan}, a standard LLM agent without explicit
planning or persistent memory;
(2)~\textbf{WebAnchor}, which improves a frozen policy through
test-time search over sampled candidate
plans~\citep{yu2026webanchoranchoringagentplanning,miromindteam2026mirothinker17h1heavyduty};
(3)~\textbf{TodoEvolve}, which trains a meta-planner to
generate task-specific planning architectures but accumulates
no experience~\citep{liu2026todoevolvelearningarchitectagent};
and (4)~\textbf{A-MapReduce}, the closest baseline to ours,
which distills reusable insights from trajectories but updates
its memory directly from task
outcomes~\citep{chen2026amapreduceexecutingwidesearch}.
All methods use the same benchmark runtimes, tools, and
verifiers.

\paragraph{Benchmarks.}
We evaluate CHIME on four long-horizon agent benchmarks:
(1)~\textbf{$\tau^2$-bench}, multi-turn customer-service
interactions~\citep{barres2025tau2benchevaluatingconversationalagents};
(2)~\textbf{VitaBench}, multi-turn life-service tasks with
changing user intent and large tool
spaces~\citep{he2025vitabenchbenchmarkingllmagents};
(3)~\textbf{BrowseComp-ZH}, long-horizon InfoSeeking~\citep{zhou2025browsecompzhbenchmarkingwebbrowsing};
and (4)~\textbf{BFCL-v4}, function calling, using
\texttt{Live} and long-horizon \texttt{Agentic}
groups~\citep{patil2025bfcl}.

\paragraph{Backbones.}
We evaluate CHIME with two foundation model backbones:
Qwen3.5-Flash model (35B total, 3B active)~\citep{qwen3.5} and
DeepSeek-V4-Flash model (284B total, 13B active)~\citep{deepseekai2026deepseekv4highlyefficientmilliontoken}.
The same backbone serves all roles in CHIME, including the
planner, the executor, and the Credit Attribution Gate.

\paragraph{Test-Time Evaluation Protocol.}
Following previous works~\citep{ye2026umemunifiedmemoryextraction,ouyang2026reasoningbankscalingagentselfevolving},
we process each benchmark as a sequential task stream and split
it 7:3 into a train split and an eval split along the stream
order. For benchmarks with multiple domains, we truncate each domain to the smallest domain size and split each domain independently, so that all domains contribute equally.
The train split measures accumulation: memory-based methods
update their memory from outcomes on this stream, and train
performance reflects the utility of the accumulated experience
for subsequent tasks.
The eval split measures generalization: we freeze the memory
banks accumulated on the train split and solve each eval task
by directly retrieving the accumulated memory, without any
further memory update, so eval performance reflects transfer
of the accumulated memory to unseen tasks rather than further
adaptation.
To reduce variance from sample ordering, we run the protocol
under three random orderings and report Avg@3 performance.

\paragraph{Implementation Details.}
All evaluation-side LLM roles use Qwen3.6-Flash~\citep{qwen36_35b_a3b} under the official benchmark protocols~\citep{barres2025tau2benchevaluatingconversationalagents,
he2025vitabenchbenchmarkingllmagents,
zhou2025browsecompzhbenchmarkingwebbrowsing}.
Memory and $\tau^2$-bench knowledge retrieval use
Qwen3-Embedding-0.6B~\citep{zhang2025qwen3embeddingadvancingtext}.
For retrieval and reranking, we set $(N,K)=(20,3)$,
$\lambda=3.0$, $\alpha=1.0$, and $\beta=0.2$.
For memory evolution, we use $\theta_{\mathrm{conf}}=0.5$,
$\theta_{\mathrm{merge}}=0.88$, $n_{\min}=3$, and
$\theta_{\mathrm{prune}}=-0.1$.
All LLM calls use temperature $0.0$, except WebAnchor plan generation ($0.6$).
Further settings are provided in Appendix.

\section{Main Experimental Results}

We analyze the main results (Table~\ref{tab:main_results}) in
three parts: memory accumulation on the train split, memory
transfer to unseen tasks on the eval split, and ablations of
CHIME's key designs.

\paragraph{Memory Accumulation (Train Split).}
CHIME achieves the highest average with both backbones (35.02\% on Qwen3.5-Flash and 38.00\% on
DeepSeek-V4-Flash), outperforming the strongest baseline by
3.38\% and 0.90\%.
These gains show that CHIME's accumulated memory guides
subsequent tasks more effectively than baseline memory.

\paragraph{Memory Transfer (Eval Split).}
With memory banks frozen, CHIME achieves the highest eval average
with both backbones: it outperforms the strongest baseline by
2.96\% on Qwen3.5-Flash and 3.68\% on DeepSeek-V4-Flash.
Compared with A-MapReduce, the closest baseline to ours,
CHIME leads by 5.37\% on Qwen3.5-Flash and 3.68\% on
DeepSeek-V4-Flash.
These gains show that attributed memory transfers better
than outcome-based memory.

\paragraph{Ablation Studies.}
We group the ablations into two parts: hierarchical memory
design, and credit attribution and memory evolution.
Table~\ref{tab:ablation_results} reports the results with
Qwen3.5-Flash:
(1)~\textbf{Hierarchical memory design:}
On the eval split, removing planning memory, execution memory,
or both reduces the average performance by 3.26\%, 4.81\%, and 3.13\%, respectively. Moreover, collapsing the two banks into a shared one (w/o Hierarchical Bank) reduces it by 3.60\%.
These results show that CHIME's gains stem from stage-specific experience and its explicit separation, rather than from memory accumulation alone.
(2)~\textbf{Credit attribution and memory evolution:}
Removing the Credit Attribution Gate (w/o Attribution Gate)
reverts memory updates to direct outcome feedback and reduces
the eval average by 2.44\%. Besides, disabling value-based
reranking (w/o Credit Rerank) and credit-aware memory evolution
(w/o Memory Evolution) reduces it by 2.04\% and 3.04\%,
respectively. These results indicate that attribution decides which stage receives feedback, while value-based reranking and credit-aware evolution decide which memory is reused and retained.

\section{Analysis}
\label{sec:analysis}

This section investigates the sources of CHIME's
effectiveness through six research questions:
memory quality (RQ1), evolution efficiency (RQ2),
memory value utility (RQ3), credit attribution gate reliability (RQ4), gains from stronger credit models (RQ5), and cross-backbone transfer (RQ6).

\paragraph{RQ1: Does CHIME Accumulate High-Quality Memory?}
We use GLM-5.2 as an external judge to evaluate every memory
in the Qwen3.5-Flash train-split banks.
Each memory receives three binary judgments:
\emph{Correctness} (valid stage-specific guidance),
\emph{Non-Redundancy} (no duplicated experience), and
\emph{Generality} (applicability beyond the source episode).
Table~\ref{tab:memory_quality} shows that CHIME achieves the
highest pass rate on all three dimensions, with 81.04\% of
its memories passing all of them.
Removing the Gate floods the bank with noisy outcome-level
traces, dropping non-redundancy and generality below 31\%;
A-MapReduce's insights reach only 36.08\% correctness, as
they are induced without credited trajectories.
These results show that credit-aware construction yields
valid, distinct, and reusable memories, supporting the
attribute-before-memorize principle.

\begin{table}[H]
\centering
\small
\renewcommand{\arraystretch}{1.08}
\begin{tabular*}{\columnwidth}{@{\extracolsep{\fill}}lcccc@{}}
\toprule
\textbf{Method}
& \textbf{Corr.}
& \textbf{Non-R.}
& \textbf{Gen.}
& \textbf{All} \\
\midrule
\textbf{CHIME}
& \textbf{81.20} & \textbf{93.60} & \textbf{90.13} & \textbf{81.04} \\
A-MapReduce
& 36.08 & 81.65 & 75.32 & 36.08 \\
w/o Attribution Gate
& 37.09 & 25.77 & 30.53 & 21.38 \\
\bottomrule
\end{tabular*}
\caption{Memory quality on the Qwen3.5-Flash train split under GLM-5.2 evaluation (\%). Corr., Non-Red., and Gen. denote correctness, non-redundancy, and generality, respectively; All denotes the fraction passing all three criteria.}
\label{tab:memory_quality}
\end{table}

\begin{figure}[t]
    \centering
    \includegraphics[width=\columnwidth]{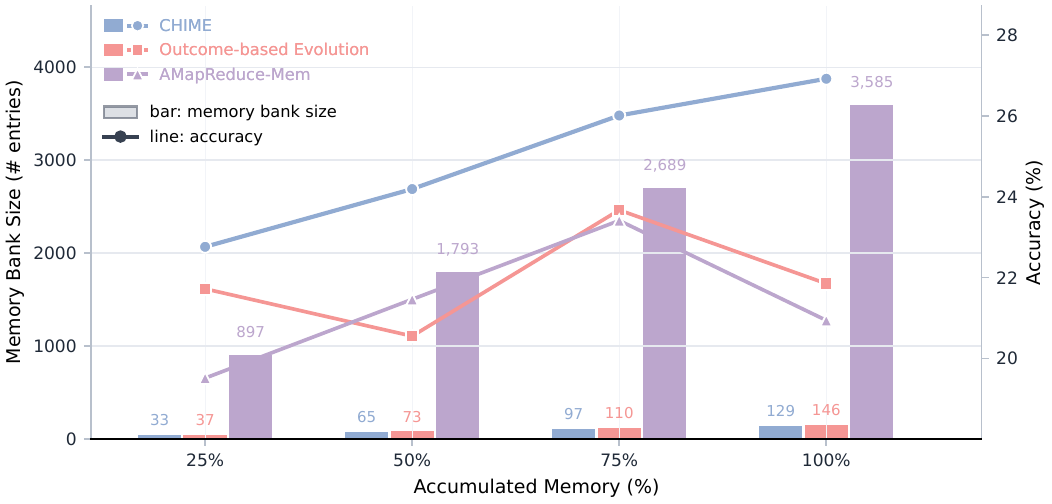}
    \caption{Memory evolution efficiency on $\tau^2$-bench. Bars show memory-bank size, and lines show task accuracy at four accumulation checkpoints.}
    \label{fig:memory_evolution_efficiency}
\end{figure}

\paragraph{RQ2: Does Stage-Attributed Evolution Improve Memory Efficiency?}
We track accuracy against memory-bank size on $\tau^2$-bench
along the train stream.
We freeze the memory bank at four accumulation checkpoints
(25\%, 50\%, 75\%, and 100\% progress of the stream) for
measurement.
We compare CHIME with two outcome-based baselines:
Outcome-based Evolution (w/o Attribution
Gate)~\cite{ye2026umemunifiedmemoryextraction} and A-MapReduce, which update
memory directly from final outcomes.
As shown in
Figure~\ref{fig:memory_evolution_efficiency}, CHIME improves
accuracy as accumulation proceeds while keeping the smallest
bank.
At the end of the stream, CHIME retains only 129 memories, compared with 3,585 for A-MapReduce,
yet achieves the highest accuracy.
In contrast, both counterparts keep growing their banks, but
their accuracy peaks at the 75\% checkpoint and declines at
full accumulation:
without attribution, misleading memories are written into the
bank and accumulate, which increasingly disturbs retrieval.
These results show that effective self-evolving memory depends
on the quality of the accumulated memories, not their quantity.

\begin{figure}[t]
    \centering
    \includegraphics[width=\columnwidth]{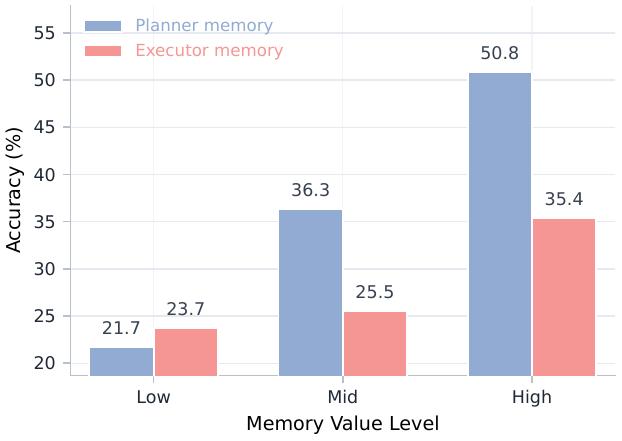}
    \caption{Downstream task accuracy by memory value level. Bars report accuracy for planning and execution memories in the low-, mid-, and high-value groups.}
    \label{fig:value_calibration}
\end{figure}

\paragraph{RQ3: Do Learned Memory Values Reflect Downstream Utility?}
We partition the retrieved planning and execution memories
into low-, mid-, and high-value groups by evenly splitting
their value range, and report the average task accuracy of
each group.
Figure~\ref{fig:value_calibration} shows a monotonic increase
at both stages: from low to high value, accuracy rises from
21.7\% to 50.8\% for planning memories and from 23.7\%
to 35.4\% for execution memories.
This correlation suggests that the learned values faithfully
reflect memory quality, indicating that the Credit Attribution
Gate attributes credit effectively.
Notably, higher-value planning memories bring larger accuracy
gains than their execution counterparts (29.1\% vs.\ 11.7\%
between the low- and high-value groups), whereas
low-value planning
memories are even less useful than low-value execution ones
(21.7\% vs.\ 23.7\%), underscoring the importance of
planning in long-horizon tasks.
Together with the value-reranking ablation in
Table~\ref{tab:ablation_results}, these results support
selecting reusable experience by learned values.

\paragraph{RQ4: Is the Credit Attribution Gate Reliable?}
We assess the reliability of the Credit Attribution Gate on
Qwen3.5-Flash along two dimensions:
\emph{Stability}, the agreement between the original
attribution and the majority label of three samples;
and \emph{Rescued}, the fraction of failed episodes whose
attributed failure no longer persists when the task is rerun
with only the generated memory.
As shown in Table~\ref{tab:gate_attribution_reliability},
planning and execution labels reach an agreement between
87.8\% and 97.9\% on both benchmarks, and between
56.3\% and 67.6\% of the failures are rescued,
indicating that the Gate assigns repeatable credit and that
the generated experience effectively corrects the attributed
failure.

\begin{table}[H]
\centering
\small
\renewcommand{\arraystretch}{1.08}
\begin{tabular*}{\columnwidth}{@{\extracolsep{\fill}}llcc@{}}
\toprule
\textbf{Benchmark}
& \textbf{Stage}
& \textbf{Stability}
& \textbf{Rescued} \\
\midrule
VitaBench & \emph{planning}  & 91.7 & 67.6 \\
          & \emph{execution} & 97.9 & 60.2 \\
\midrule
BFCL-v4   & \emph{planning}  & 87.8 & 64.7 \\
          & \emph{execution} & 92.9 & 56.3 \\
\bottomrule
\end{tabular*}
\caption{Gate attribution reliability (\%) on VitaBench and BFCL-v4.
\emph{Stability}: agreement between the original attribution and the majority label of three samples.
\emph{Rescued}: fraction of failed episodes whose attributed failure no longer occurs when the task is rerun with only the generated memory.}
\label{tab:gate_attribution_reliability}
\end{table}

\paragraph{RQ5: Does Stronger Attribution Further Improve
Performance?}
The Credit Attribution Gate relies on the backbone's self-reflection capability.
We replace it with a stronger model, Qwen3.7-Flash, for both
credit attribution and experience generation, and evaluate
the resulting memory on the eval split.
Table~\ref{tab:stronger_credit} shows consistent gains on
all three benchmarks, from 2.25\% to 3.12\%.
These gains indicate that more accurate attribution yields
more effective experience, so CHIME benefits directly from
stronger credit models.
\begin{table}[H]
\centering
\small
\renewcommand{\arraystretch}{1.08}
\begin{tabular*}{\columnwidth}{@{\extracolsep{\fill}}lccc@{}}
\toprule
\textbf{Benchmark}
& \textbf{Self-Reflect}
& \textbf{Qwen3.7-Flash}
& \textbf{$\Delta$} \\
\midrule
$\tau^2$-Bench & 23.80 & \textbf{26.92} & +3.12 \\
VitaBench      & 12.50 & \textbf{15.00} & +2.50 \\
BFCL-v4        & 72.45 & \textbf{74.70} & +2.25 \\
\bottomrule
\end{tabular*}
\caption{Eval accuracy (\%). Qwen3.5-Flash model is replaced by stronger Qwen3.7-Flash for credit attribution gate. Results are from a single run rather than Avg@3.}
\label{tab:stronger_credit}
\end{table}

\paragraph{RQ6: Does CHIME's Memory Transfer Across
Backbones?}
We copy the memory accumulated by a source backbone on a
benchmark's train split into a target backbone's frozen
eval run.
As shown in Table~\ref{tab:cross_model_transfer}, w/o Mem.\ denotes the target running without memory, and Self denotes the
target with its self-accumulated memory.
CHIME's transferred memory outperforms A-MapReduce's in all
four settings by between 2.25\% and 4.68\%, and closely
approaches the upper bound: matching or exceeding it on the
Qwen3.5-Flash target model, and staying within 1.72\% on BFCL-v4 for the DeepSeek-V4-Flash target model.
These experimental results reveal that attributed memory thus encodes reusable guidance rather than backbone-specific patterns.

\begin{table}[H]
\centering
\small
\renewcommand{\arraystretch}{1.1}
\begin{tabular*}{\columnwidth}{@{\extracolsep{\fill}}lcccc@{}}
\toprule
\textbf{Benchmark} & \textbf{w/o Mem.} & \textbf{A-Map.} & \textbf{CHIME} & \textbf{Self} \\
\midrule
\multicolumn{5}{@{}l}{\emph{DeepSeek-V4-Flash $\to$ Qwen3.5-Flash}} \\
$\tau^2$-Bench & 24.45 & 21.72 & \textbf{26.40} & 25.92 \\
BFCL-v4        & 70.86 & 71.66 & \textbf{73.91} & \textbf{73.91} \\
\midrule
\multicolumn{5}{@{}l}{\emph{Qwen3.5-Flash $\to$ DeepSeek-V4-Flash}} \\
$\tau^2$-Bench & 30.04 & 28.22 & 30.82 & \textbf{36.28} \\
BFCL-v4        & 60.13 & 61.85 & 65.43 & \textbf{67.15} \\
\bottomrule
\end{tabular*}
\caption{Cross-model memory transfer on the eval split (\%). Each group header denotes the source $\to$ target backbones.}
\label{tab:cross_model_transfer}
\end{table}

\section{Conclusion}
\label{sec:conclusion}

In this work, we presented CHIME, an self-evolving agentic planning framework based on our proposed attribute-before-memorize principle, which separates and updates planning and execution memory bank through stage-level credit attribution.
Experiments on four long-horizon agent benchmarks show that CHIME consistently outperforms strong training-based and self-evolving memory baselines.
Further analyses validate CHIME along six dimensions: memory quality, evolution efficiency, memory value utility, credit attribution gate reliability, credit-model scaling, and cross-backbone transfer.
These results suggest that reliable memory evolution depends
not only on what agents remember, but also on whether each
experience is attributed to the decision stage where it can
provide effective guidance.

\clearpage
\appendix

\section{Implementation Details}
\label{app:implementation_details}
\begin{table*}[t]
\centering
\small
\renewcommand{\arraystretch}{1.08}
\begin{tabular}{@{}p{0.18\textwidth}p{0.28\textwidth}p{0.48\textwidth}@{}}
\toprule
\textbf{Component} & \textbf{Setting} & \textbf{Value} \\
\midrule
\textit{Models and decoding}
& Agent backbone
& \texttt{qwen3.5-flash} or \texttt{deepseek-v4-flash} \\
& Evaluation-side LLM
& \texttt{qwen3.6-flash} \\
& Default decoding
& Temperature $0.0$; thinking disabled \\
& WebAnchor plan generation
& Four candidates; temperature $0.6$ \\
& Maximum input/output length
& $245{,}000$ / $16{,}384$ tokens \\
& Request policy
& $2{,}400$-s timeout; at most 15 retries; 50-s base retry interval \\
\midrule
\textit{Retrieval and reranking}
& Embedding configuration
& \texttt{Qwen3-Embedding-0.6B}; batch size 16; maximum length $16{,}384$ \\
& Candidate/retrieved set sizes
& $(N,K)=(20,3)$ per memory layer \\
& Candidate similarity threshold
& $0.3$ \\
& Reranking parameters
& $\lambda=3.0$, $\alpha=1.0$, and $\beta=0.2$ \\
& Near-tie relevance margin
& $0.03$ \\
\midrule
\textit{Memory evolution}
& Update confidence threshold
& $\theta_{\mathrm{conf}}=0.5$ \\
& Experience-addition threshold
& $0.1$ \\
& Initial memory state
& $(v_i,n_i)=(0,0)$ \\
& Merge threshold
& $\theta_{\mathrm{merge}}=0.88$ \\
& Pruning thresholds
& $n_{\min}=3$ and $\theta_{\mathrm{prune}}=-0.1$ \\
& Memory capacity
& Unlimited \\
\midrule
\textit{Evaluation protocol}
& Train/eval split
& $7{:}3$ within each benchmark domain \\
& Selected memory checkpoint
& Epoch 2 for $\tau^2$-Bench, VitaBench, and BFCL-v4; epoch 5 for BrowseComp-ZH \\
\midrule
\textit{$\tau^2$-Bench}
& User-simulator LLM
& \texttt{qwen3.6-flash} \\
& Interaction controls
& 100 steps; at most 10 errors; seed 42 \\
& Subprocess timeout
& $3{,}600$ s \\
& Knowledge retrieval
& Local embedding retrieval; Top-$8$ \\
\midrule
\textit{VitaBench}
& User-simulator LLM
& \texttt{qwen3.6-flash} \\
& Agent/user configuration
& \detokenize{llm_agent}/\detokenize{user_simulator} \\
& Benchmark language
& English \\
& Interaction controls
& 100 steps; at most 10 errors; seed 42 \\
\midrule
\textit{BrowseComp-ZH}
& Search framework
& \texttt{smolagents} \\
& Search-agent backbone
& \texttt{qwen3.5-flash} or \texttt{deepseek-v4-flash}, matching the evaluated backbone \\
& Search configuration
& Top-$10$; at most 10 search steps \\
& Search/page-fetch timeout
& 120 s each \\
& Page-length limit
& $40{,}000$ characters \\
& Answer judge
& Deterministic \texttt{qwen3.6-flash} \\
\midrule
\textit{BFCL-v4}
& Evaluator
& \texttt{qwen3.6-flash} checker \\
& Search configuration
& Top-$10$; at most 10 tool/search steps \\
\midrule
\textit{Shared execution}
& Episode concurrency
& 16 for $\tau^2$-Bench, BrowseComp-ZH, and BFCL-v4; 8 for VitaBench \\
\bottomrule
\end{tabular}
\caption{Implementation parameters for the main experiments. Benchmark-specific settings are shared across methods.}
\label{tab:implementation_parameters}
\end{table*}

CHIME keeps the agent backbone frozen throughout, updating only the external planning and execution memory banks during accumulation and freezing them during transfer. These updates use only interaction trajectories and environment feedback, not reference answers. For consistent comparison, all methods share the same benchmark setup and evaluation-side models, and each baseline uses the best-performing configuration reported in its paper.
All results are averaged over three random instance orderings (Avg@3).

We evaluate CHIME with \texttt{Qwen3.5-Flash}~\citep{qwen3.5} and
\texttt{Deepseek-V4-Flash}~\citep{deepseekai2026deepseekv4highlyefficientmilliontoken};
in each setting, the same backbone handles
planning, execution, and self-reflection in the Credit Attribution Gate.
All evaluation-side LLM roles use
\texttt{Qwen3.6-Flash}~\citep{qwen36_35b_a3b}, while
\texttt{Qwen3-Embedding-0.6B}~\citep{zhang2025qwen3embeddingadvancingtext}
serves as the retriever.
Table~\ref{tab:implementation_parameters} lists the complete implementation
settings.

\section{Failure Analysis}
\label{app:error_analysis_boundaries}

Our analysis reveals two practical boundaries of CHIME.

\paragraph{Agent Capability.}
CHIME augments the agent with memory guidance but leaves the backbone unchanged.
As a result, even useful memory may not improve the outcome when the agent cannot follow the guidance or execute the required actions correctly.
In the gate reliability analysis, rerunning failed episodes with the generated memory resolves 56.3--67.6\% of attributed failures, but not all of them.
The benefit of CHIME therefore remains bounded by the underlying capabilities of the agent.

\paragraph{Memory Transferability.}
Our cross-backbone transfer results show that memory can transfer across backbones when the benchmark remains unchanged.
Tasks within the same benchmark generally share the conditions under which a memory applies, including the available tools, valid actions, and success criteria.
In contrast, we observe weaker transfer across benchmarks, where semantic similarity does not ensure that these conditions remain unchanged.
CHIME's memory is therefore most transferable across tasks with similar scenarios and operating conditions, while transfer across different benchmarks remains limited.

\section{Case Studies}
\label{app:case_studies}

We include three representative cases to illustrate how the memory bank is used and updated. The first case shows a planner-side benefit, where retrieved memories help decompose a complex request before execution. The second case shows an executor-side benefit, where retrieved memories help bind exact tool arguments. The third case shows the attribution gate itself: a failed trajectory is routed to the planner layer because the executor followed the plan and the reusable error lies in the planning abstraction.

\begin{figure*}[t]
\centering
\begin{tcolorbox}[
    colback=gray!5!white,
    colframe=black!70!white,
    title=\textbf{Case Study: Planner Memory for Multi-Status Retail Requests},
    fonttitle=\bfseries\small,
    boxrule=0.8pt,
    arc=3mm,
    left=5pt, right=5pt, top=5pt, bottom=5pt
]
\small

\textbf{Benchmark:} $\tau^2$ Bench \\
\textbf{Scores:} Ours: $1.0$ \,$\mid$\, AMapReduce-Mem: $0.0$ \,$\mid$\, NoPlan: $0.0$ \\
\rule{\linewidth}{0.4pt}
\vspace{2pt}

\textbf{Task Input:} \\
The user wants to return the bookshelf and jigsaw puzzle received in the same delivered order; return the backpack from another delivered order that also contains a vacuum cleaner; modify a pending order by changing its shipping address to the user's default Chicago address and changing the item color to red; and obtain the tracking number of a cancelled order. The user reveals information gradually and refers to different orders by item names rather than order IDs.
\vspace{2pt}

\textbf{Retrieved Planner Memories:}
\begin{itemize}
    \setlength{\itemsep}{2pt}
    \setlength{\parskip}{0pt}
    \item \textbf{Item-status mapping.} For multi-item retail requests, explicitly map each item to its order ID and status before choosing an action. Pending orders should use \detokenize{modify_pending_order_items}, while delivered orders should use \detokenize{return_delivered_order_items}.
    \item \textbf{Cancelled-order handling.} Before planning to reverse or operate on a cancelled order, first verify whether a supported tool exists; do not assume cancelled state changes are reversible.
    \item \textbf{One-shot pending modification.} For pending-order modifications, collect all requested item/address changes into one call because the order may not allow incremental follow-up modifications.
\end{itemize}

\textbf{Ours Trajectory:}
\begin{itemize}
    \setlength{\itemsep}{2pt}
    \setlength{\parskip}{0pt}
    \item \textbf{Step 1:} Authenticate Lucas Brown and retrieve his order history.
    \item \textbf{Step 2:} Separate the request into four status-specific subproblems: delivered-order return for bookshelf+jigsaw, delivered-order return for backpack, pending-order modification for address+color, and cancelled-order tracking lookup.
    \item \textbf{Step 3:} Return the complete bookshelf+jigsaw item set from order \detokenize{#W6239298} and return the backpack from order \detokenize{#W9218746}.
    \item \textbf{Step 4:} Modify pending order \detokenize{#W4860251} once, bundling the Chicago address update and red-item replacement into the same pending-order modification.
    \item \textbf{Step 5:} Retrieve and report the tracking number \texttt{286422338955} for the cancelled order \detokenize{#W1154986}.
\end{itemize}

\textbf{Prediction:} \\
Ours reports that it processed the two delivered-order returns, modified the pending order's address and color, and provided the tracking number \texttt{286422338955} for the cancelled order. The official Tau2 verifier marks the episode correct.
\vspace{2pt}

\textbf{Baseline Contrast:} \\
AMapReduce-Mem produces a plausible plan, but its final answer reports the tracking number of the returned bookshelf/jigsaw order rather than treating the cancelled order as a distinct status. NoPlan also performs several local actions but fails the official verifier. The difference is that Ours uses planner memory to form the correct order-status decomposition before execution.

\end{tcolorbox}
\caption{Planner memory helps the agent decompose a single natural-language request into the correct status-conditioned retail actions. The selected memories are useful before tool execution because they determine which objects should be grouped together and which tool family applies to each group.}
\label{fig:case_planner_memory_tau2_retail}
\end{figure*}

\begin{figure*}[t]
\centering
\begin{tcolorbox}[
    colback=gray!5!white,
    colframe=black!70!white,
    title=\textbf{Case Study: Executor Memory for Account-Level Net Credit},
    fonttitle=\bfseries\small,
    boxrule=0.8pt,
    arc=3mm,
    left=5pt, right=5pt, top=5pt, bottom=5pt
]
\small

\textbf{Benchmark:} $\tau^2$ Bench \\
\textbf{Scores:} Ours: $1.0$ \,$\mid$\, AMapReduce-Mem: $0.0$ \,$\mid$\, NoPlan: $0.0$ \\
\rule{\linewidth}{0.4pt}
\vspace{2pt}

\textbf{Task Input:} \\
Kim Junho has three checking accounts with nine ATM-fee discrepancies. The agent must verify the user, retrieve transaction histories for the Blue, Green, and Light Green accounts, cross-reference account-specific ATM-fee rules, identify both overcharges and missing fees, and apply the correct net credits: Blue Account \detokenize{$9.50}, Green Account \detokenize{$9.00}, and Light Green Account \detokenize{$1.50}.
\vspace{2pt}

\textbf{Retrieved Executor Memory:} \\
When applying net credits to multiple bank accounts with mixed overcharges and undercharges, sum all overcharge refunds, sum all missing or undercharged fees, subtract the undercharges from the overcharges, and verify the final account-level amount before invoking the credit tool. Do not apply a single credit amount based only on overcharges or only on undercharges.
\vspace{2pt}

\textbf{Ours Trajectory:}
\begin{itemize}
    \setlength{\itemsep}{2pt}
    \setlength{\parskip}{0pt}
    \item \textbf{Step 1:} Verify the user and unlock the account and transaction-history tools.
    \item \textbf{Step 2:} Retrieve transaction histories for all three checking accounts: \detokenize{chk_kj93a7b2e1_1}, \detokenize{chk_kj93a7b2e1_2}, and \detokenize{chk_kj93a7b2e1_3}.
    \item \textbf{Step 3:} Compute account-level net credits rather than treating each discrepancy as an isolated dispute.
    \item \textbf{Step 4:} Invoke \detokenize{apply_checking_account_credit_5829} with three account-specific amounts: \detokenize{$9.50}, \detokenize{$9.00}, and \detokenize{$1.50}.
    \item \textbf{Step 5:} Confirm that all discrepancies were corrected and that no further user action is needed.
\end{itemize}

\textbf{Prediction:} \\
Ours summarizes the final credits for all three accounts and the official Tau2 database check is correct. The final state matches the required account-level net-credit updates.
\vspace{2pt}

\textbf{Baseline Contrast:} \\
NoPlan collapses the task to two disputes and reports only \detokenize{$4.50} in total credits, missing most required account-level corrections. AMapReduce-Mem also fails the official verifier. The useful distinction is execution-level: the plan can say ``review all accounts and compute credits,'' but reward depends on binding the exact account IDs and net amounts in the final tool calls.

\end{tcolorbox}
\caption{Executor memory helps when the high-level plan is conceptually correct but success depends on precise action arguments. Here the memory guides account-level arithmetic and tool-argument binding, preventing the agent from collapsing several discrepancies into an incomplete dispute summary.}
\label{fig:case_executor_memory_tau2_banking}
\end{figure*}

\begin{figure*}[t]
\centering
\begin{tcolorbox}[
    colback=gray!5!white,
    colframe=black!70!white,
    title=\textbf{Case Study: Gate Attribution for a Strict Deadline Failure},
    fonttitle=\bfseries\small,
    boxrule=0.8pt,
    arc=3mm,
    left=5pt, right=5pt, top=5pt, bottom=5pt
]
\small

\textbf{Benchmark:} VitaBench \\
\textbf{Outcome:} Failed episode, verifier score $0.0$ with partial rubric score $0.75$ \\
\textbf{Gate Label:} \detokenize{learn_layer=planner} \,$\mid$\, confidence $0.90$ \,$\mid$\, \detokenize{external_issue=false} \\
\rule{\linewidth}{0.4pt}
\vspace{2pt}

\textbf{Task Input:} \\
The user wants to order specific Sichuan dishes from a previously visited restaurant to a temporary work location. The food should arrive before a friend's lunch break starts at 13:00, and it must respect low-oil and low-salt dietary constraints.
\vspace{2pt}

\textbf{Agent Plan:} \\
The plan correctly selects Xiao Sichuan (Shifan Street Branch), verifies the temporary work address Jinzheng Haiyue International, filters the menu for Boiling Fish and garlic-flavored vegetables, and schedules delivery around the lunch deadline.
\vspace{2pt}

\textbf{Final Prediction / Verifier Feedback:} \\
The final answer confirms that the order is set, but the official evaluator finds one critical violation: the estimated delivery time is \texttt{2024-09-12 13:00:00}. The rubric requires delivery strictly before 13:00, so exactly 13:00 does not satisfy the constraint.
\vspace{2pt}

\textbf{Gate Diagnosis:}
\begin{itemize}
    \setlength{\itemsep}{2pt}
    \setlength{\parskip}{0pt}
    \item \textbf{Plan sufficient:} \texttt{false}. The plan handled store and item selection, but failed to encode a strict-inequality time buffer.
    \item \textbf{Execution followed plan:} \texttt{true}. The executor carried out the planned schedule; the problem was not a wrong tool argument relative to the plan.
    \item \textbf{External issue:} \texttt{false}. The failure is reusable and not caused by an unavailable API or environment error.
    \item \textbf{Routed layer:} \texttt{planner}. The missing lesson concerns how to plan around strict ``before'' deadlines.
\end{itemize}

\textbf{Memory Written by the Gate:} \\
For delivery orders that require arrival \emph{before} a specific deadline, calculate the dispatch time by subtracting the estimated shipping duration plus a safety buffer from the deadline, and verify that the resulting delivery time is strictly less than the deadline. Avoid setting a delivery time that lands exactly on the boundary.

\end{tcolorbox}
\caption{The gate converts a failed trajectory into a layer-specific memory. Because the executor followed the plan and the remaining error was a strict temporal-constraint mistake, the episode writes a planner memory rather than polluting the executor memory bank.}
\label{fig:case_gate_attribution_vitabench}
\end{figure*}

\paragraph{Planner-side decomposition.}
Figure~\ref{fig:case_planner_memory_tau2_retail} shows why planner memory cannot be replaced by a generic summary of past successes. The task contains several retail operations that look similar in natural language but require different protocols: returned items belong to delivered orders, color and address edits belong to a pending order, and the requested tracking number belongs to a cancelled order. The retrieved planner memories directly encode this object-status decomposition. As a result, the agent plans the correct action family for each object before making tool calls, while the baselines produce plausible but verifier-failing summaries.

\paragraph{Executor-side precision.}
Figure~\ref{fig:case_executor_memory_tau2_banking} highlights a different failure mode. The high-level plan is not the main challenge: all methods can state that the agent should review accounts and correct ATM-fee errors. The reward depends on execution details: retaining three account identities, computing net credits after subtracting missing fees, and invoking the credit tool with the exact amount for each account. The executor memory is useful because it is written at the same granularity as the eventual action arguments.

\paragraph{Layer-aware attribution.}
Figure~\ref{fig:case_gate_attribution_vitabench} illustrates why the gate uses trajectory-level diagnosis rather than only the binary verifier outcome. The delivery episode fails, but it should not penalize executor memory: the selected restaurant, address, and item choices are mostly correct, and the executor follows the planned schedule. The reusable error is the planner's treatment of a strict ``before'' constraint as if equality were acceptable. The gate therefore writes a planner memory about strict deadline buffers and leaves the executor bank untouched. This keeps the memory bank more specific and reduces cross-layer contamination.

\end{document}